\documentclass[letterpaper, 10 pt, conference]{ieeeconf}  

\usepackage{url}
\usepackage{graphicx}
\usepackage{booktabs}
\usepackage{romannum}
\usepackage{amsfonts}
\usepackage{multirow}
\usepackage[table,xcdraw]{xcolor}
\usepackage{array}
\usepackage{float}
\usepackage[export]{adjustbox} 
\usepackage{algorithm}
\usepackage{balance}
\usepackage{cite}
\usepackage{algpseudocode} 
\usepackage{longtable}
\usepackage{booktabs}

\usepackage{graphics}
\usepackage{upgreek}
\usepackage{amssymb}
\usepackage{refstyle}

\usepackage{amsmath}
\usepackage{subfigure}
\usepackage{todonotes}
\usepackage{mathtools}
\usepackage[para,online,flushleft]{threeparttable}

\usepackage{hyperref}

\usepackage{cleveref}

\newcommand\copyrighttext{%
  \footnotesize \textcopyright 2024 IEEE. Personal use of this material is permitted.
  Permission from IEEE must be obtained for all other uses, in any current or future
  media, including reprinting/republishing this material for advertising or promotional
  purposes, creating new collective works, for resale or redistribution to servers or
  lists, or reuse of any copyrighted component of this work in other works.
}
\newcommand\copyrightnotice{%
\begin{tikzpicture}[remember picture,overlay]
\node[anchor=south,yshift=10pt] at (current page.south) {\fbox{\parbox{\dimexpr\textwidth-\fboxsep-\fboxrule\relax}{\copyrighttext}}};
\end{tikzpicture}%
}

\IEEEoverridecommandlockouts                              

\title{\LARGE \bf
Close the Sim2real Gap via Physically-based Structured Light Synthetic Data Simulation}
\author{Kaixin Bai$^{1,2}$, Lei Zhang$^{1,2}$, Zhaopeng Chen$^{2*}$, Fang Wan$^{3*}$, Jianwei Zhang$^{1}$
\thanks{*Corresponding authors.}
\thanks{{$^{1}$TAMS (Technical Aspects of Multimodal Systems), Department of
Informatics, Universit\"at Hamburg, Germany}, {$^{2}$Agile Robots AG, Munich, Germany} , {$^{3}$School of Design, Southern University of Science and Technology, Shenzhen, Guangdong, China} }
}%
\DeclarePairedDelimiterX{\norm}[1]{\lVert}{\rVert}{#1}

\begin{document}

\maketitle
\copyrightnotice

\thispagestyle{empty}
\pagestyle{empty}

\begin{abstract}    

Despite the substantial progress in deep learning, its adoption in industrial robotics projects remains limited, primarily due to challenges in data acquisition and labeling. Previous sim2real approaches using domain randomization require extensive scene and model optimization. To address these issues, we introduce an innovative physically-based structured light simulation system, generating both RGB and physically realistic depth images, surpassing previous dataset generation tools. We create an RGBD dataset tailored for robotic industrial grasping scenarios and evaluate it across various tasks, including object detection, instance segmentation, and embedding sim2real visual perception in industrial robotic grasping. By reducing the sim2real gap and enhancing deep learning training, we facilitate the application of deep learning models in industrial settings.
Project details are available at \href{https://baikaixin-public.github.io/structured_light_3D_synthesizer/}{https://baikaixin-public.github.io/structured\_light\_3D\_synthesizer/}.

\end{abstract}

\section{Introduction}
\label{intro}
Data collection for computer and robotic vision tasks, particularly for object segmentation and 6D pose annotation~\cite{marion2018label}, is labor-intensive and challenging. Additionally, gathering industrial data for deep learning models can be problematic due to factory rules, confidentiality, and safety concerns.

To address the challenges in obtaining real-world data, sim2real methods have been proposed to generate synthetic RGB images in 3D simulators for tasks such as robotic perception~\cite{morrical2021nvisii, josifovski2018object, ummadisingu2022cluttered, arents2022synthetic}, autonomous driving~\cite{pollok2019unrealgt,khan2019procsy}, intelligent agriculture solutions~\cite{toda2020training, barth2018data}, consumer and manufacturing~\cite{ummadisingu2022cluttered, wong2019synthetic}, and medical treatment~\cite{cartucho2020visionblender}, to reduce manual labor and improve the performance of deep learning models. Domain randomization~\cite{tobin2017domain} has been employed to generate photorealistic RGB images. This approach minimizes the sim2real gap by altering lighting, object material, and texture, but demands expert rendering knowledge and significant optimization within the 3D simulator.

While various tools have been created to generate photo-realistic simulation datasets, they are constrained by the performance and capabilities of their respective simulation engines~\cite{morrical2021nvisii}. Differing strengths between game engines and film industry renderers, as well as inconsistencies like the left-handed coordinate system, present challenges for tasks like object pose estimation and robotic applications.

Decreasing prices of RGBD cameras have increased their use for computer and robot vision tasks, particularly for 3D visual perception and semantic information of the environment. This has led to an increase in using depth images as inputs to neural networks or in combination with RGB images~\cite{dxiang2020learning, huang2021survey}, to improve the performance of vision tasks. Additionally, RGBD images are used as inputs for multimodal perception tasks~\cite{xu2017multi, wang2019densefusion}. Depth images have been used as inputs for neural networks to train robots for perception and grasping tasks~\cite{zakharov2018keep, danielczuk2018segmenting}. Researchers have attempted to reduce the gap between real and simulated depth images by generating physically-based depth images using stereo cameras or TOF cameras in virtual environments~\cite{planche2017depthsynth,gschwandtner2011blensor}, or applying post-processing using neural networks such as GANs~\cite{zakharov2018keep,yuan2022sim} to further align them with real ones.

Robotic tasks in industrial settings require visual recognition capabilities for diverse objects in terms of location, placement direction, type, shape, and size. These tasks often require identifying a large number of objects for grasping in cluttered scenes with objects of different sizes, which typically requires matching object instance segmentation or object detection to localize the objects and pose estimation based on point cloud or texture analysis to determine the grasping pose with high accuracy. 
Structure light cameras are widely used in industries like automotive, and logistics. They can provide 2D and 3D information with high precision and are adaptable to various requirements such as anti-ambient light, high accuracy, high reconstruction speed, and small size.

\begin{figure*}[!h]
	\begin{center}
		\includegraphics[width=17.6cm]{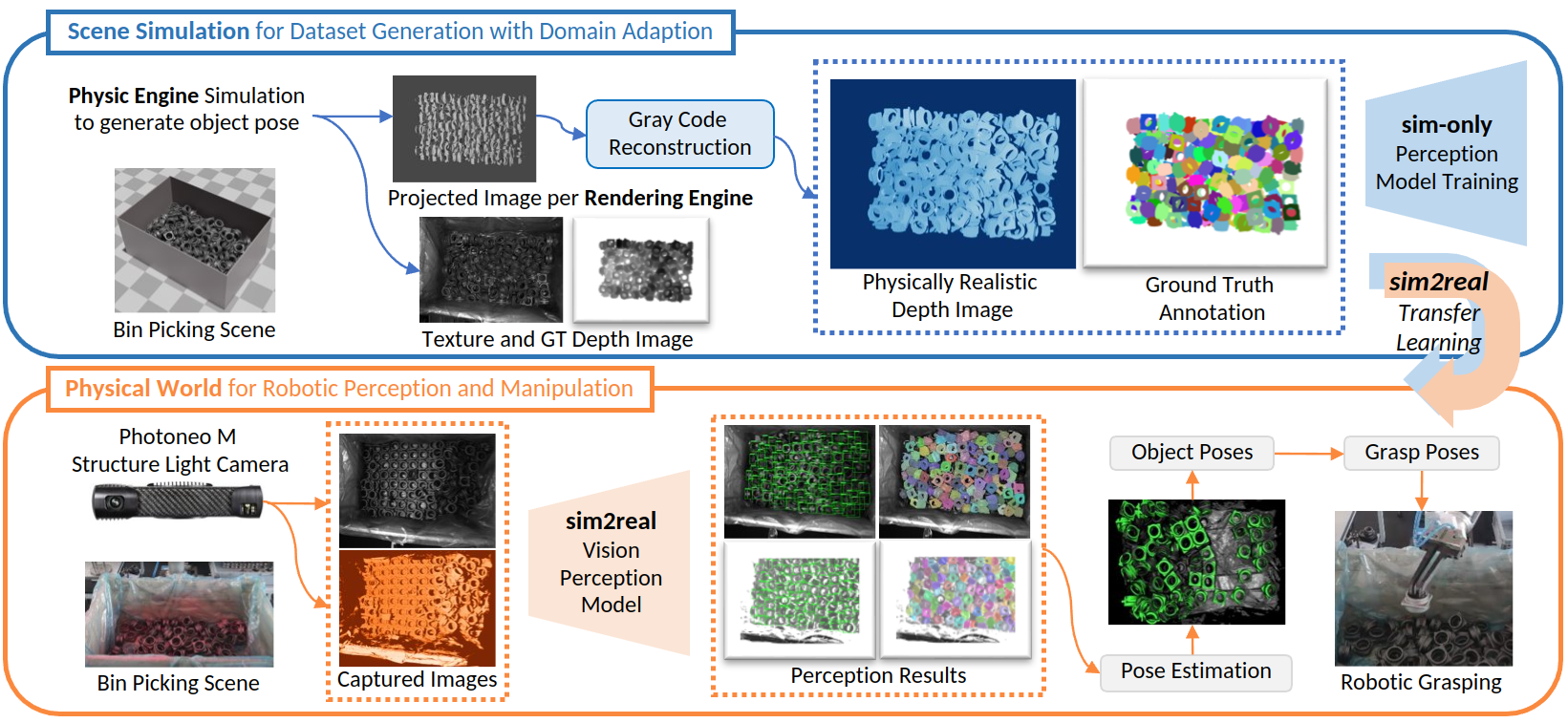}
		\caption{Pipeline of physically-based sim2real transfer learning: We built a physically-based simulator with gravity to generate realistic data of cluttered scenarios. Then we use ray tracing to perform structured light projection, then decode and reconstruct the projected image to render physically-based realistic depth images with ground truth annotations for instance segmentation. Next, we train the vision perception using simulation only and perform sim2real transfer learning on the trained model to ensure good perception results in real-world scenarios. After that, we apply pose estimation to obtain the poses of objects and perform model-based robotic grasping. }
		\label{fig.simulation_arch_part1}
	\end{center}
\end{figure*}

We propose a data generator for physically-based gray code structure light camera simulation. This generates photorealistic RGB and physically-based synthetic depth data, complete with 3D reconstruction algorithm noises, for robotic sim2real tasks. Our key contributions are:
\begin{itemize}
\item A physically-based gray code structured light camera simulation data generator, built using the Blender Cycles rendering engine and Optix AI denoiser, which generates photorealistic RGB data, physically-based synthetic depth data, and ground truth annotations for object pose, 2D/3D bounding boxes, and segmentation.
\item A dataset with physically-realistic simulated RGBD data as a training set and real data as a test set, which can be utilized to evaluate the sim2real performance gap and generalization ability of vision perception tasks like object detection and instance segmentation.
\item We provide a real-world demonstration of the effectiveness of our sim2real data generation and robot perception network based on this data generation method in actual robot tasks.
\end{itemize}

\section{Related Work}\label{sec:relatedwork}

\subsection{Synthetic Dataset Generation}

The trend of training robots on synthetic datasets and transferring to real-world datasets is gaining traction. Various simulation data generation tools and plug-ins have emerged, ranging from game engines like Omniverse~\cite{rojas2022easy} and Unreal Engine 4~\cite{qiu2017unrealcv,to2018ndds,khan2019procsy}, to 3D simulation tools like Blender~\cite{heindl2021blendtorch,arents2022synthetic,mata2022standardsim,im2019deep} and PyBullet~\cite{josifovski2018object, ummadisingu2022cluttered}. These tools vary in supported programming languages, headless rendering capabilities, and real-time rendering performance. While game engines often excel in frame rates using raster-based methods, they may lack accurate light transport simulation due to the absence of ray tracing, which limits their rendering performance for reflective and transparent objects.

\subsubsection{\textbf{Sim2real Gap}}

The sim2real gap remains a major hurdle for deep learning methods trained on synthetic datasets for vision tasks and robotic manipulation. This gap arises due to disparities between synthetic RGB data and real-world conditions, influenced by environmental and camera parameters. Minimizing this gap often requires extensive optimization of object materials, lighting, and sensor characteristics. To tackle this, researchers employ domain randomization techniques to vary colors, lighting, and noise~\cite{to2018ndds,tobin2017domain}, and domain adaptation methods to address data domain mismatches, particularly in GANs training~\cite{yuan2022sim,bousmalis2017unsupervised}.

\subsection{Physically-based Synthetic Depth Sensor Simulation}

Industrial 3D cameras are commonly classified into passive stereo, active stereo, ToF, and structured light types. Physically-based sensor simulation is crucial for improving the quality of datasets for vision and robotic tasks. Depth images are increasingly favored in perception tasks due to their lower sim2real gaps. For example, Danielczuk et al.\cite{xiang2021learning} explored object segmentation methods using depth images and non-photorealistic RGB images for feature learning.

Various methods have been developed to narrow the sim2real gap in depth images. For instance, Zakharov et al.\cite{zakharov2018keep} and Danielczuk et al.\cite{danielczuk2018segmenting} utilize ground truth depth images to train robots for perception tasks. Others like Planche et al.\cite{planche2017depthsynth} and Gschwandtner et al.\cite{gschwandtner2011blensor} generate synthetic depth images using virtual stereo or TOF cameras. To further align synthetic and real depth images, Zakharov et al.\cite{zakharov2018keep} and Yuan et al.\cite{yuan2022sim} employ neural networks like GANs for post-processing. The significance of depth image simulation across different 3D cameras is thus evident.

\begin{figure*}[!h]
	\centering
	\includegraphics[width=17.7cm]{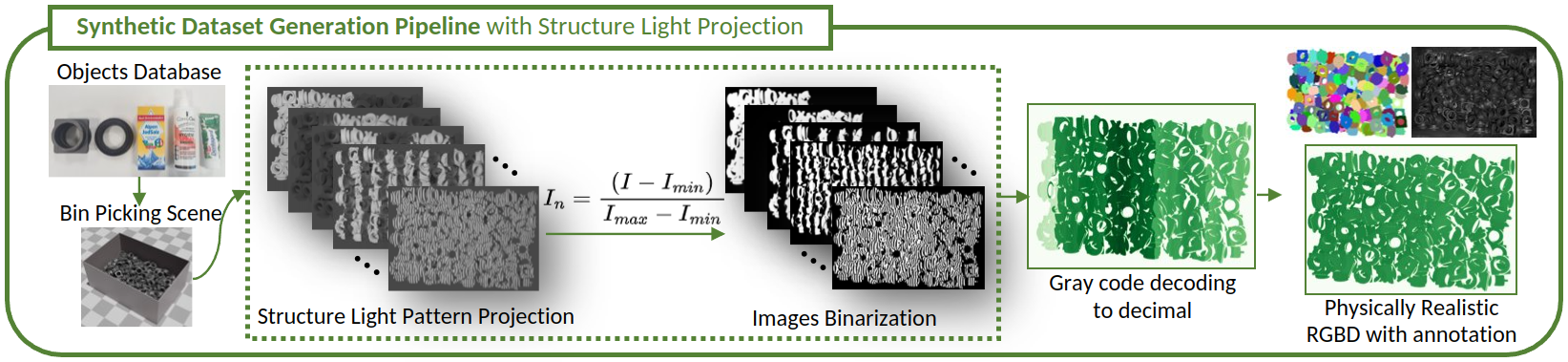}
	\caption{The system architecture of the proposed pattern projection simulation and 3D reconstruction methods is as follows: First, we generate a scene using the proposed physically-based rendering technique. Then, we project various gray code patterns into the scene. These patterns serve as binary images for 3D reconstruction. The system includes a structured-light camera setup consisting of one projector and one camera. This setup captures the projected patterns and the scene. Finally, we estimate a synthetic depth image with realistic structured light noise using our proposed methods.}
	\label{fig.simulation_arch_part2}
\end{figure*}

\section{Method}\label{sec:method}

To close the simulation-reality gap for gray code structured light cameras, we've created a data simulator using Blender. Using physically-based rendering, we simulate scenes and apply gray code patterns. Realistic depth images are then generated through 3D reconstruction. Our system employs NVIDIA's OptiX engine and GeForce RTX 3070 Ti GPU for rendering and includes OptiX's AI-accelerated denoiser~\cite{parker2010optix} to enhance image render speed. Fig.~\ref{fig.simulation_arch_part1} outlines the physically-based rendering pipeline for robotic pick-and-place tasks, while Fig.~\ref{fig.simulation_arch_part2} shows pattern projection and 3D reconstruction. We focus on ray tracing's acknowledged benefits, without comparing it to other rendering techniques like rasterization rendering.

\subsection{Physically-Based Rendering}\label{sec:rendering}

In our study, we employ physically-based rendering to simulate real-world scenes featuring a wooden box with densely arranged objects. This involves dividing the space above the box into voxel grids, sampling these grids based on object count, and then dropping the objects into the box while considering realistic physics like collisions. The final renders include synthetic RGB and ground-truth depth images, as well as instance segmentation. Real-world RGBD images and point clouds from an empty wooden box are incorporated for domain adaptation.

Structured light cameras often suffer from decoding errors due to varying light paths caused by material properties and lighting conditions. This introduces noise into depth maps. To capture this realistically, our simulator uses ray tracing for gray code pattern projection, enabling accurate depth maps with noise characteristics similar to real-world applications. This approach bypasses the shortcomings of rasterization-based methods, which neglect light path calculations, resulting in less realistic images.

\subsection{Pattern Projection}\label{sec:rendering}
To simulate the rendering process based on the physical principles of structured light cameras in the real world, we present a projector based on a spotlight with textured light in Blender to simulate the pattern projection of structured light cameras. During pattern projection, the gray code pattern images are separately projected onto the top of the object scene using our proposed projector. The corresponding gray images are then rendered, as shown in Fig.~\ref{fig.simulation_arch_part2}.

The rendering of high-quality images is typically time-consuming in the film industry. In synthetic dataset generation, using the same workflow as film rendering is not cost-effective for vision tasks and robotic manipulation. In recent years, AI algorithms have been applied to reduce the rendering time for high-fidelity images~\cite{chaitanya2017interactive}. For example, the AI denoiser with Optix integration in Blender can speed up rendering for pixels with a limited number of samples. It is even possible to achieve real-time rendering after just two frames while maintaining the quality of the rendered images~\cite{zhang2022close}. To speed up our dataset generation, we use the AI denoiser to render both RGB images and projected pattern images after rendering 20 frames. The parameter is chosen based on our qualitative comparison experiment.

\subsection{3D Reconstruction with Gray Code Pattern}\label{sec:3D_reconstruction}
To reconstruct the scene from images rendered with gray code pattern projection, we first generate binary images, as illustrated in Fig.~\ref{fig.simulation_arch_part2}. Each point within the projected area experiences at least one brightness shift, achieved by incorporating both fully black and fully white images. For every pixel, we compute its temporal maximum and minimum values to establish a binarization threshold. A pixel is assigned a value of 1 if the threshold exceeds 0.5, and 0 otherwise. 
We then use 3D reconstruction structured-light techniques, as described in\cite{li2017high}, to reconstruct the scene. In this process, the projector is modeled as a pinhole camera. We obtain decoding images based on the gray code encoding, and the object's imaging position on the camera plane in pixels is represented by $(u_c, v_c)$. The virtual imaging position under the projector camera model is indicated by $(u_p,v_p)$. Using this information, we can model the structured-light system with one projector and one camera using the following formula:
\begin{center}
\begin{equation}
s_{c}\begin{bmatrix}
u_{c}\\ 
v_{c}\\ 
1
\end{bmatrix}=K_{c}[R_{c}|t_{c}]\begin{bmatrix}
X\\ 
Y\\ 
Z\\ 
1
\end{bmatrix},
s_{p}\begin{bmatrix}
u_{p}\\ 
v_{p}\\ 
1
\end{bmatrix}=K_{p}[R_{p}|t_{p}]\begin{bmatrix}
X\\ 
Y\\ 
Z\\ 
1
\end{bmatrix}
\end{equation}
\end{center}

where $K_c$ and $K_p$ are the intrinsic parameters of the camera and the projector. When the camera coordinate system coincides with the world coordinate system, rotation matrix $R_{c}$ can be formulated as unit matrix and translation vector $t_{c}$ of camera can be formulated as all zero matrix. 
By decoding the Gray code we can obtain the correspondence of each pixel under the camera and projector model. Finally, we can reconstruct the 3D scene with the following equation:
\begin{figure}[htbp]
	\begin{center}
		\includegraphics[width=6cm]{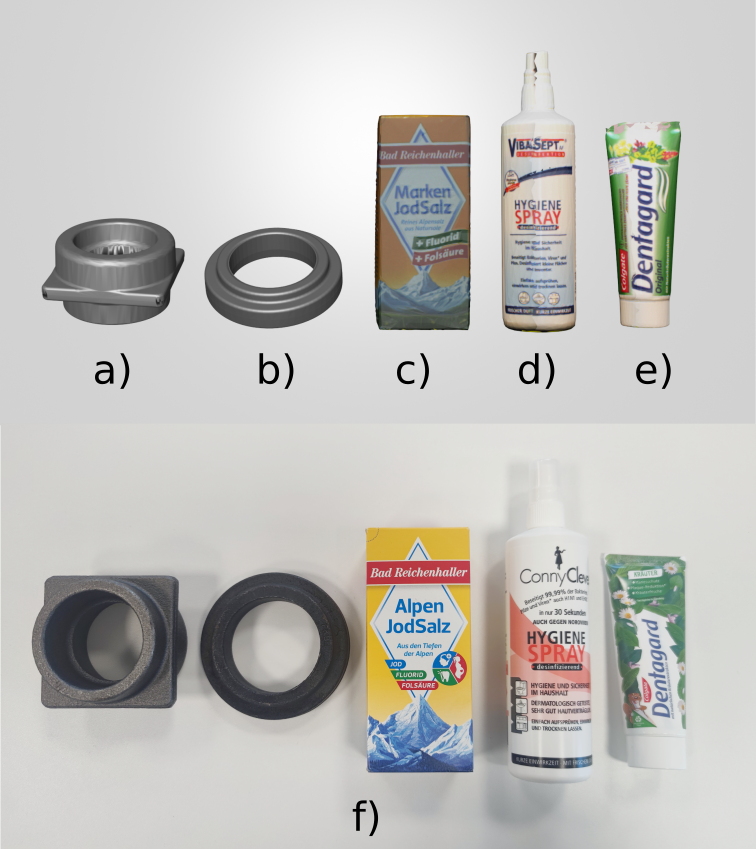}
		\caption{Object Database. a) Metal Workpiece 1. b) Metal Circle Workpiece 2. c)YellowSaltCube (KIT object models database). d) HygieneSpray (KIT object models database). e) Toothpaste (KIT object models database). f) The corresponding real objects.}
		\label{fig.simulation_arch}
	\end{center}
\end{figure}
\begin{gather}
\begin{split}
M_{c}=&K_{c}[R_{c}\mid t_{c}]=\begin{bmatrix}
m_{11}^{c} & m_{12}^{c}  & m_{13}^{c} & m_{14}^{c}\\ 
m_{21}^{c} & m_{22}^{c} & m_{23}^{c} & m_{24}^{c}\\ 
m_{31}^{c} & m_{32}^{c} & m_{33}^{c} & m_{34}^{c}
\end{bmatrix} \\
M_{p}=&K_{p}[R_{p}\mid t_{p}]=\begin{bmatrix}
m_{11}^{p} & m_{12}^{p}  & m_{13}^{p} & m_{14}^{p}\\ 
m_{21}^{p} & m_{22}^{p} & m_{23}^{p} & m_{24}^{p}\\ 
m_{31}^{p} & m_{32}^{p} & m_{33}^{p} & m_{34}^{p}
\end{bmatrix}  \\
&\begin{bmatrix}
X\\ 
Y\\ 
Z
\end{bmatrix}=M\begin{bmatrix}
m_{34}^{c}u_{c}-m_{14}^{c}\\ 
m_{34}^{c}v_{c}-m_{24}^{c}\\ 
m_{34}^{p}u_{p}-m_{14}^{p}
\end{bmatrix} \\
M=&\begin{bmatrix}
m_{11}^{c}-m_{31}^{c}u_{c} &m_{12}^{c}-m_{32}^{c}u_{c}  &m_{13}^{c}-m_{33}^{c}u_{c} \\ 
m_{21}^{c}-m_{31}^{c}v_{c} &m_{22}^{c}-m_{32}^{c}v_{c}  &m_{23}^{c}-m_{33}^{c}v_{c} \\ 
m_{11}^{p}-m_{31}^{p}u_{p} &m_{12}^{p}-m_{32}^{p}v_{c}  &m_{13}^{p}-m_{33}^{p}u_{p} 
\end{bmatrix}^{-1}
\end{split}
\end{gather}

\section{Experiments}\label{sec:experiments}

To validate the efficacy of our synthetic data simulator designed for gray code structured light cameras, we assess its performance on tasks like instance segmentation and object detection. For this, we employ a synthetic training dataset and a real-world testing dataset. Initially, we introduce an object database specifically crafted for generating cluttered environments. Subsequently, we produce an extensive, photorealistic dataset featuring single-class, multi-instance scenes, suitable for tasks such as instance segmentation and object detection. A real-world dataset, aligned with our object database, serves as the testing set.

\subsection{Object Database}\label{sec:object_database}

In our object database, we have gathered two metal parts in a dark gray color from industry, as well as three household objects from the KIT object models database~\cite{kasper2012kit}. These objects encompass common items found in pick and place scenarios, including industrial components and supermarket merchandise frequently encountered in robotic tasks.

\subsection{Synthetic and Real-world Datasets}\label{sec:synthetic_and_real_world_datasets}
To explore vision tasks and robotic operations in industrial settings, we create an extensive synthetic dataset featuring single-class, multi-instance scenes from our object database. This dataset comprises rendered RGB and depth images, synthetic depth images using gray code structured light reconstruction, along with ground truths for object poses, 2D bounding boxes, and instance segmentation masks. For the objects in our database, we employ Run-Length Encoding (RLE) for ground truth representation in synthetic data and use polygon encoding for real-world data to streamline manual annotations.
\begin{table*}[htbp]
\centering
\vspace*{1mm}
	\caption{Quantitative evaluation results of object detection and instance segmentation.}
	\label{tab.train_results}
	\vspace*{-3mm}
\begin{threeparttable}
\begin{tabular}{|cc|ccc|ccc|}
\hline
\multicolumn{2}{|c|}{Tasks} & \multicolumn{3}{c|}{Object Detection (YOLOv3)$^{*}$} & \multicolumn{3}{c|}{Instance Segmentation (SOLOv2)$^{*}$} \\ \hline
\multicolumn{1}{|l|}{\multirow{2}{*}{Object}} & \multicolumn{1}{l|}{\multirow{2}{*}{Input Type}}                                        &
\multicolumn{2}{l|}{Test Set Type$^{**}$}   & \multicolumn{1}{l|}{\multirow{2}{*}{Sim2real Gap$^{***}$}} & \multicolumn{2}{l|}{Test Set Type$^{**}$}   & \multicolumn{1}{l|}{\multirow{2}{*}{Sim2real Gap$^{***}$}} \\ \cline{3-4} \cline{6-7} 
\multicolumn{1}{|l|}{}                                          & 
\multicolumn{1}{l|}{}                                        &
\multicolumn{1}{l|}{Synthetic}   & \multicolumn{1}{l|}{Real}  & {} & \multicolumn{1}{l|}{Synthetic}   & \multicolumn{1}{l|}{Real}  & {} \\ \hline
\multicolumn{1}{|l|}{\multirow{2}{*}{Metal Workpiece 1.}}       & 
RGB & \multicolumn{1}{l|}{0.624} & \multicolumn{1}{l|}{0.547} & 0.077       & \multicolumn{1}{l|}{0.711} & \multicolumn{1}{l|}{0.620} & 0.091       \\ \cline{2-8} 
\multicolumn{1}{|l|}{}                                & 
Ground Truth Depth   & \multicolumn{1}{l|}{0.613} & \multicolumn{1}{l|}{0.527} & 0.086       & \multicolumn{1}{l|}{0.692} & \multicolumn{1}{l|}{0.557} & 0.135       \\ \cline{2-8} 
\multicolumn{1}{|l|}{}                                & 
Proposed Synthetic Depth   & \multicolumn{1}{l|}{0.634} & \multicolumn{1}{l|}{\textbf{0.617}} & 0.017       & \multicolumn{1}{l|}{0.698} & \multicolumn{1}{l|}{\textbf{0.646}} & 0.052       \\ \hline
\multicolumn{1}{|l|}{\multirow{3}{*}{Metal Circle Workpiece 2.}}           & 
RGB & \multicolumn{1}{l|}{0.673} & \multicolumn{1}{l|}{0.614} & 0.059       & \multicolumn{1}{l|}{0.646} & \multicolumn{1}{l|}{0.565} & 0.081        \\ \cline{2-8} 
\multicolumn{1}{|l|}{}                                & 
Ground Truth Depth   & \multicolumn{1}{l|}{0.662} & \multicolumn{1}{l|}{0.633} & 0.029       & \multicolumn{1}{l|}{0.636} & \multicolumn{1}{l|}{0.544} & 0.092       \\ \cline{2-8} 
\multicolumn{1}{|l|}{}                                & 
Proposed Synthetic Depth   & \multicolumn{1}{l|}{0.674} & \multicolumn{1}{l|}{\textbf{0.628}} & 0.046       & \multicolumn{1}{l|}{0.634} & \multicolumn{1}{l|}{\textbf{0.627}} & 0.007       \\ \hline
\multicolumn{1}{|l|}{\multirow{2}{*}{Toothpaste}}     & 
RGB & \multicolumn{1}{l|}{1.000} & \multicolumn{1}{l|}{0.885} & 0.115        & \multicolumn{1}{l|}{1.000} & \multicolumn{1}{l|}{0.953} & 0.047        \\ \cline{2-8} 
\multicolumn{1}{|l|}{}                                & 
Ground Truth Depth   & \multicolumn{1}{l|}{0.999} & \multicolumn{1}{l|}{0.950} & 0.049       & \multicolumn{1}{l|}{1.000} & \multicolumn{1}{l|}{0.970} & 0.03       \\ \cline{2-8} 
\multicolumn{1}{|l|}{}                                & 
Proposed Synthetic Depth   & \multicolumn{1}{l|}{1.000} & \multicolumn{1}{l|}{\textbf{0.980}} & 0.020        & \multicolumn{1}{l|}{1.000} & \multicolumn{1}{l|}{\textbf{0.979}} & 0.021        \\ \hline
\multicolumn{1}{|l|}{\multirow{2}{*}{YellowSaltCube}} & 
RGB & \multicolumn{1}{l|}{1.000} & \multicolumn{1}{l|}{0.826} & 0.174        & \multicolumn{1}{l|}{1.000} & \multicolumn{1}{l|}{0.621} & 0.379        \\ \cline{2-8} 
\multicolumn{1}{|l|}{}                                & 
Ground Truth Depth   & \multicolumn{1}{l|}{0.996} & \multicolumn{1}{l|}{0.678} & 0.318       & \multicolumn{1}{l|}{0.998} & \multicolumn{1}{l|}{0.802} & 0.196       \\ \cline{2-8} 
\multicolumn{1}{|l|}{}                                & 
Proposed Synthetic Depth   & \multicolumn{1}{l|}{1.000} & \multicolumn{1}{l|}{\textbf{0.827}} & 0.173         & \multicolumn{1}{l|}{1.000} & \multicolumn{1}{l|}{\textbf{0.863}} & 0.137        \\ \hline
\multicolumn{1}{|l|}{\multirow{2}{*}{HygieneSpray}}   & 
RGB & \multicolumn{1}{l|}{1.000} & \multicolumn{1}{l|}{0.947} & 0.053        & \multicolumn{1}{l|}{1.000} & \multicolumn{1}{l|}{0.963} & 0.037        \\ \cline{2-8} 
\multicolumn{1}{|l|}{}                                & 
Ground Truth Depth   & \multicolumn{1}{l|}{0.999} & \multicolumn{1}{l|}{0.797} & 0.202       & \multicolumn{1}{l|}{1.000} & \multicolumn{1}{l|}{0.875} & 0.125       \\ \cline{2-8} 
\multicolumn{1}{|l|}{}                                & 
Proposed Synthetic Depth   & \multicolumn{1}{l|}{0.980} & \multicolumn{1}{l|}{\textbf{0.953}} & 0.027        & \multicolumn{1}{l|}{1.000} & \multicolumn{1}{l|}{\textbf{0.966}} & 0.034        \\ \hline
\end{tabular}
 \begin{tablenotes}
        \footnotesize
        \item[*] Average precision (AP) @[ Intersection over Union (IoU)=0.50 ]  \\
        \item[**] Synthetic: Evaluation with synthetic dataset. Real: Evaluation with real-world dataset.\\
        \item[***] Sim2real Gap: Difference of evaluation results in different types of test set.
      \end{tablenotes}
    \end{threeparttable}
    \vspace*{-3mm}
\end{table*}
\begin{figure*}[htbp]
	\begin{center}
		\includegraphics[width=15cm]{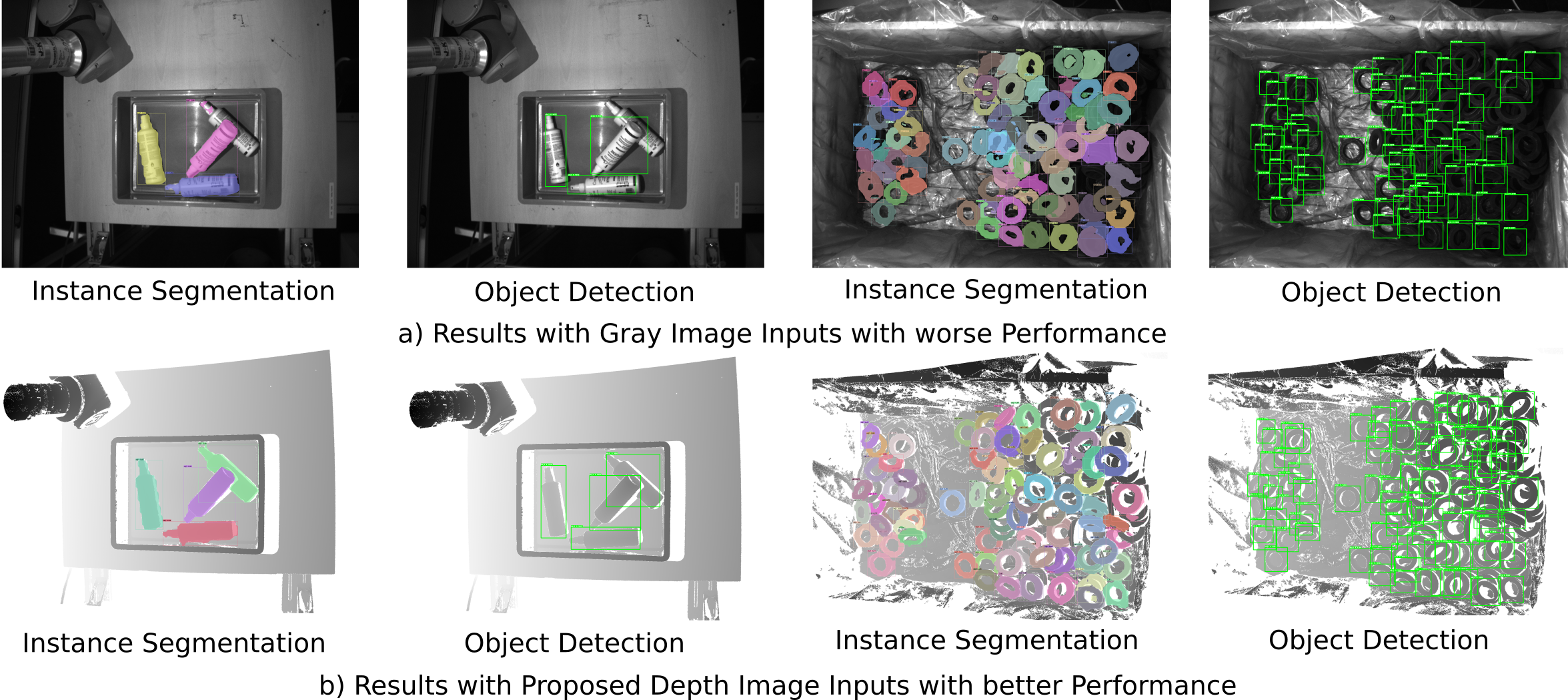}
		\caption{Qualitative Visualization results of visual perception in real-world dataset using trained models of our synthetic datasets. The depth image-based deep learning model has better performance than the model with texture image as input.}
		\label{fig.04_segmentation_result_in_real_dataset}
	\end{center}
\end{figure*}
In our physics-based rendering simulation, each scene is a single class with multiple instances. For the two types of metal parts, there are a maximum of 255 objects in the physics simulation and rendering in each scene, and for objects in the KIT dataset, there are a maximum of 10 objects in each bin box.
In the synthetic domain adaption dataset, we generate 1,000 scenes for each object in the database after simulating pattern projection and 3D reconstruction. For each scene we generate colored image, ground truth depth image, our proposed depth image and instance segmentation map. For the real-world dataset, we collect 100 sets of data for each object as the test set. In addition, we generated a domain randomization dataset using Isaac Sim~\cite{rojas2022easy} for comparison.

\subsection{Object Detection and Instance Segmentation Experiments}\label{sec:Model training}

To validate our synthetic data pipeline, we separately examine object detection and instance segmentation tasks using both real and synthetic RGB and depth images. We benchmark using YOLOv3 and SOLOv2 for these tasks, and test on real-world datasets. Quantitative results are summarized in Table~\ref{tab.train_results}. We opted for YOLOv3 due to its lack of data augmentation like Mosaic used in later versions, ensuring fairness in validating dataset efficacy. 

\begin{figure}[htbp]
	\begin{center}
		\includegraphics[width=8cm]{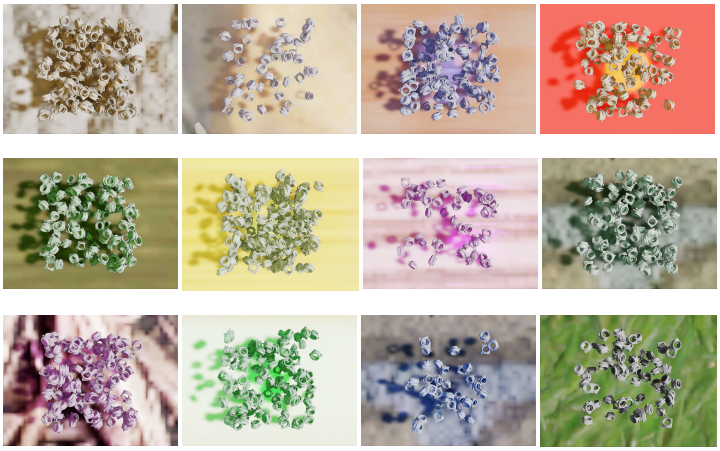}
		\caption{The creation of a domain-randomized dataset using the Isaac Sim platform. The dataset generation features varied scene backgrounds, object numbers and poses, and lighting conditions in terms of intensity and color. Each parameter is randomized to increase the diversity of the dataset.}
		\label{fig.domain_random}
	\end{center}
\end{figure}

To gauge the real-world efficacy of our domain-adapted dataset, we employ YOLOv7 to compare performance against a domain-randomized dataset in perception tasks. YOLOv7's data augmentation features align well with industrial scenarios. The randomized dataset, created with Isaac Sim, varies scene backgrounds, object quantities and poses, and lighting as shown in Fig.~\ref{fig.domain_random}.

\subsection{Robotic grasping experiment}\label{sec:Robotic grasping experiment}

\begin{figure}[htbp]
	\begin{center}
		\includegraphics[width=8cm]{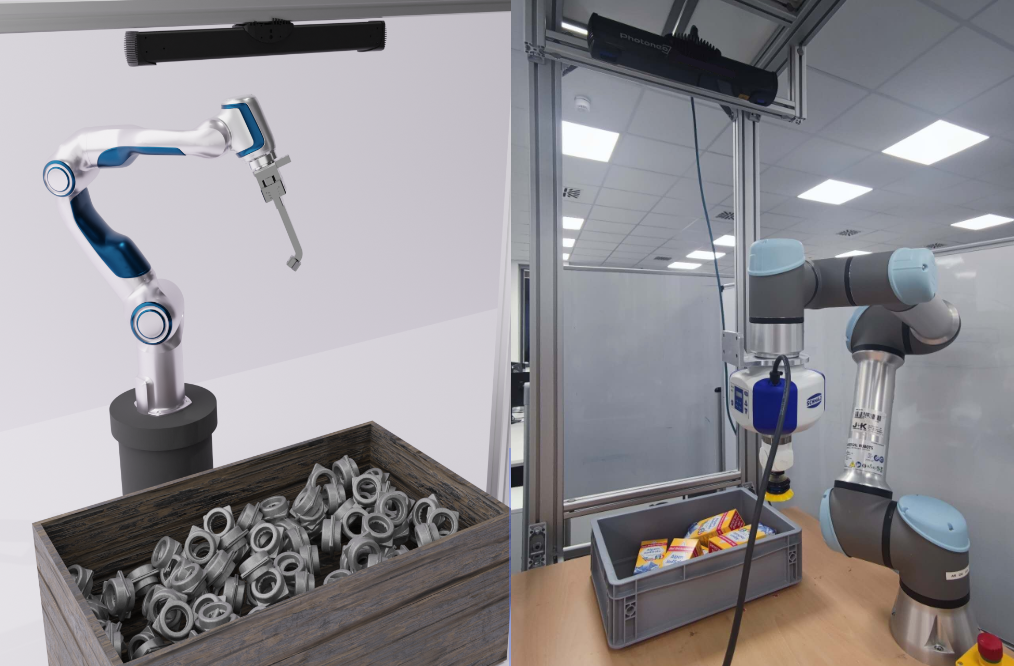}
		\caption{Comparative experimental setups showcasing robotic grasping. On the left, a model-based grasping scenario is depicted, utilizing a Diana7 robot and a Photoneo L camera for industrial part manipulation. On the right, a half-model based grasping approach is illustrated, employing a UR5e robot and a Photoneo M camera.}
		\label{fig.robot_setup}
	\end{center}
\end{figure}

To evaluate our vision perception model in robotic grasping experiments, we build the setup of model-based and half-model-based grasping experiments. Both setups are designed to address the sim2real performance gap. For the model-based grasping experiment, we employ the Diana7 robot and Photoneo L industrial part. Meanwhile, the half-model grasping experiment involves the UR5e robot and a Photoneo M camera as shown in Fig.~\ref{fig.robot_setup}. In our experiments, we employ the objects from our dataset to execute bin-picking tasks, adhering to the half-model-based grasping procedure as detailed in~\cite{du2021vision} and Dex-Net3.0~\cite{mahler2018dex}. The performance of visual perception has a profound influence on the computational burden of the grasping algorithm and directly affects the success rate of grasping. Our approach involves initially detecting the object in a depth image via an object detection technique, and then applying a grasping detection method. We further examine the variations in grasping success rate and manipulation speed following the integration of our visual perception approach. Both setups are devised with the aim of demonstrating that the use of sim2real in visual perception tasks can significantly enhance the performance metrics, such as success rate and algorithm runtime, in robotic grasping applications.

\section{Results}\label{sec:results}
\subsection{Qualitative Study}

\begin{figure}[htbp]
	\begin{center}
		\includegraphics[width=7cm]{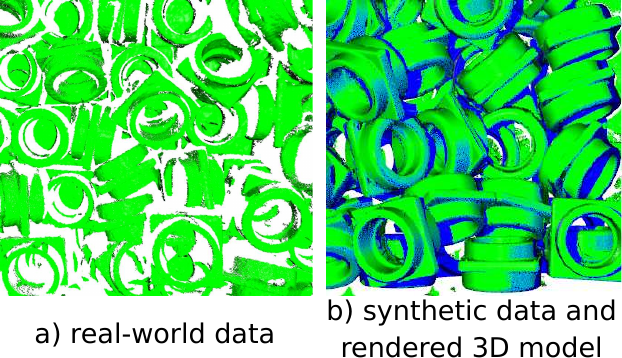}
		\caption{Qualitative Results of Structured Light Noise. a) Localized area of real-world point cloud of cluttered objects (Metal Workpiece 1.). b) Partical area of synthetic data with synthetic shadows(zero value in depth image) and sharp noise(flying noise), rendered point cloud in green and 3D model with ground truth pose in blue.}
		\label{fig.05_qualitative_results}
	\end{center}
\end{figure}

Assessing the disparity between reconstructed depth data from our synthetic data simulator and real-world scene data, we conduct a qualitative analysis via localized visualizations. Fig.~\ref{fig.05_qualitative_results} a) depicts a real-world scene's local depth image of cluttered metal workpiece 1, while Fig.~\ref{fig.05_qualitative_results} b) displays synthetic data from a structured-light camera (green) and rendered 3D model data (blue). Our generator can simulate shadow and sharp noise of structured-light cameras based on the difference between its synthetic data and 3D model renderings.

Fig.~\ref{fig.05_qualitative_results} b) further illustrates the efficacy of our proposed structured light-based data simulator. The depicted point cloud noise from our simulator closely matches the noise from a real structured-light camera, suggesting minimal data disparity between our simulated depth images and point clouds and the real counterparts. This infers that our simulator likely incurs less performance loss in visual perception tasks using depth images or point clouds.

\subsection{Quantitative Studies}
\subsubsection{Object Detection and Instance Segmentation}
Table~\ref{tab.train_results} details the quantitative results of our data simulation methods in object detection and instance segmentation tasks, with varying input types (RGB or depth images) and test datasets (synthetic or real-world).

Initially, a sim2real gap emerges when evaluating the trained model on real-world data across all objects. This gap is measured by subtracting the average precision results of intersection over union (IoU) in real-world data from the synthetic data results. This sim2real gap is positive for selected household objects, while qualitative results suggest its presence for two metal industrial objects, attributed to issues like shadows and sharp noise.

Depth images are less sensitive to lighting and appearance changes, offering robustness against environmental variations. Using depth images as input can lessen a model's computational load, often containing less information than RGB images. They allow the distinction between object shape and texture, enhancing certain perception tasks' performance. For instance, object recognition often emphasizes shape over texture, and using a depth map as input enables model focus on the object's shape. Moreover, depth images provide additional information about camera-object distances, beneficial for tasks like 3D reconstruction and robot navigation.

In addition, we conducted tests on the YOLOv7 model using datasets based on both our domain adaptation approach and domain randomization to ascertain the potential performance of our proposal in real-world projects. The experiments demonstrated that the domain adaptation-based RGB dataset achieved an IoU of 0.628 on the real dataset for the Metal Workpiece 1 object, whereas the model based on our proposed depth image achieved an IoU of 0.648. Both exceeded the IoU of 0.584 derived from the domain adaptation-based RGB dataset. The analysis suggests that in industrial grasping scenarios, which are usually static, the performance boost brought by domain adaptation surpasses that of domain randomization.

In conclusion, utilizing depth images as input in deep learning perception tasks can bolster model robustness, efficiency, and performance, providing a more durable representation of 3D structure and facilitating shape-texture separation. Moreover, they exhibit less sensitivity to lighting and appearance changes.

\subsection{Robotic Grasping Experiment}

Incorporating sim2real perception into our half-model-based grasping pipeline resulted in an uptick in successful grasping rates from 95.6\% to 98.0\% with Dex-Net3.0. The introduction of precise object detection concentrates the sampling of grasp points more on a single object, ultimately increasing the probability of grasp points being centered on the object's plane. As a result, the success rate of grasping also improves.

The sim2real approach significantly improved the model-based grasping task, which traditionally relies on error-prone and limited manual data annotation for object detection training. This method not only reduced project development time from two weeks to two days but also enhanced detection precision, thus bolstering system robustness. It mitigated issues related to oversized bounding boxes increasing pose estimation time and undersized ones compromising estimation accuracy and grasping success.

The vision perception model in our experiment, trained within a simulated environment and leveraging our proposed synthetic structured light-based depth images, performed on par in real-world scenarios, successfully completing the robotic grasping task. The design of our data generation pipeline, mindful of sensor noise generation, enables effective domain adaptation in real-world applications.

\section{Conclusion}\label{sec:conclusion}
Despite the progress in deep learning for perception, its industrial application remains limited due to the high cost and time required for data annotation and model adaptation. To address this, we introduce a sim2real data generation tool designed for 3D structured light cameras, commonly used in industrial robotics. The tool uses physics-based simulations to generate realistic depth maps and annotations, enabling efficient sim2real transfer learning with minimal performance loss. This innovation is crucial for integrating deep learning into industrial contexts.

Our quantitative analysis validates the tool's efficacy in perception tasks, highlighting that our physically realistic synthetic depth inputs accelerate domain adaptation and improve network performance. This significantly reduces the time needed for domain randomization, a common bottleneck in prior works.

Looking ahead, our roadmap includes expanding the Fraunhofer IPA Bin-Picking dataset and optimizing pose estimation and robotic grasping algorithms using our generated dataset, catering to a broader range of industrial applications.

\section*{ACKNOWLEDGMENT}
This research has received funding from the German Research Foundation (DFG) and the National Science Foundation of China (NSFC) in project Crossmodal Learning, DFG TRR-169/NSFC 61621136008, partially supported by ULTRACEPT (778602). 

%

\bibliographystyle{IEEEtran}
\bibliography{main}

\end{document}